\documentclass[sigconf]{acmart}
\usepackage{multirow}
\usepackage{geometry}
\usepackage{algorithm}
\usepackage{algorithmic}
\usepackage{amsthm}
\usepackage{subfigure}
\usepackage{verbatimbox}
\usepackage{balance}
\usepackage[misc]{ifsym}

\newtheorem{fact}{Fact}

\AtBeginDocument{%
  \providecommand\BibTeX{{%
    \normalfont B\kern-0.5em{\scshape i\kern-0.25em b}\kern-0.8em\TeX}}}

\copyrightyear{2022}
\acmYear{2022}
\setcopyright{acmcopyright}
\acmConference[WSDM '22] {Proceedings of the Fifteenth ACM International Conference on Web Search and Data Mining}{February 21--25, 2022}{Tempe, AZ, USA.}
\acmBooktitle{Proceedings of the Fifteenth ACM International Conference on Web Search and Data Mining (WSDM '22), February 21--25, 2022, Tempe, AZ, USA}
\acmPrice{15.00}
\acmISBN{978-1-4503-9132-0/22/02}
\acmDOI{10.1145/3488560.3498401}

\begin{document}
\fancyhead{}

\title{Learning-To-Ensemble by Contextual Rank Aggregation in E-Commerce}

\author{Xuesi Wang}\authornote{Both authors contributed equally to the paper.}
\email{xuesi.wxs@alibaba-inc.com}
\affiliation{
  \institution{Alibaba Group}
  \city{Hangzhou}
  \country{China}
}

\author{Guangda Huzhang}\authornotemark[1]
\email{guangda.hzgd@alibaba-inc.com}
\affiliation{
  \institution{Alibaba Group}
  \city{Hangzhou}
  \country{China}
}

\author{Qianying Lin}\authornote{Corresponding author.}
\email{qianying.lqy@alibaba-inc.com}
\affiliation{
  \institution{Alibaba Group}
  \city{Hangzhou}
  \country{China}
}

\author{Qing Da}
\email{daq@lamda.nju.edu.cn}
\affiliation{
  \institution{Alibaba Group}
  \city{Hangzhou}
  \country{China}
}

\renewcommand{\shortauthors}{Wang, Huzhang and Lin, et al.}

\begin{abstract}
Ensemble models in E-commerce combine predictions from multiple sub-models for ranking and revenue improvement. Industrial ensemble models are typically deep neural networks, following the supervised learning paradigm to infer conversion rate given inputs from sub-models. However, this process has the following two problems. Firstly, the point-wise scoring approach disregards the relationships between items and leads to homogeneous displayed results, while diversified display benefits user experience and revenue. Secondly, the learning paradigm focuses on the ranking metrics and does not directly optimize the revenue. In our work, we propose a new Learning-To-Ensemble (LTE) framework RA-EGO, which replaces the ensemble model with a contextual Rank Aggregator (RA) and explores the best weights of sub-models by the Evaluator-Generator Optimization (EGO). To achieve the best online performance, we propose a new rank aggregation algorithm TournamentGreedy as a refinement of classic rank aggregators, which also produces the best average weighted Kendall Tau Distance (KTD) amongst all the considered algorithms with quadratic time complexity. Under the assumption that the best output list should be Pareto Optimal on the KTD metric for sub-models, we show that our RA algorithm has higher efficiency and coverage in exploring the optimal weights. Combined with the idea of Bayesian Optimization and gradient descent, we solve the online contextual Black-Box Optimization task that finds the optimal weights for sub-models given a chosen RA model. RA-EGO has been deployed in our online system and has improved the revenue significantly.
\end{abstract}

\ccsdesc[500]{Applied computing~Online shopping}
\ccsdesc[500]{Mathematics of computing~Permutations and combinations}
\ccsdesc[500]{Information systems~Decision support systems}
\keywords{Rank Aggregation, Contextual Black-Box Optimization}


\maketitle

\section{Introduction}
Deep ensemble learning is a popular choice for E-commerce ensemble models. Deep ensemble learning models input sub-model scores and are trained for click-through rate or conversion rate prediction, following the supervised learning paradigm as a Learning-To-Rank (LTR) task. It has successfully improved the revenue for many online systems~\cite{zhu2020ensembled,li2020improving,guo2017deepfm,wang2017deep,ma2018entire}. 
The sub-models usually model the relationships between the target item and the current \emph{context} (i.e., query, user) differently. For example, sub-models may estimate the preferences of the user on stores, brands, prices, and sales separately.  
A well-trained deep ensemble learning model allows for joint mining of different user behavior patterns for a single task.

However, as the items are scored independently, similar items get similar scores thus their ranking positions are close. The homogeneous result is not desirable, as it usually benefits to display items with some variations. For example, displaying two identical items together unlikely results in two clicks, and replacing one of them with another dissimilar item can possibly result in user clicks. 
Additionally, the offline training objective (e.g., loss functions, ranking metrics) has a significant gap to the true business objective (e.g., number of purchases). Many practitioners report that a fine-tuned model with a better offline performance does not correspondingly improve the online performance ~\cite{huzhang2020aliexpress,rossetti2016contrasting,mcnee2006being,beel2013comparative,rohde2018recogym}. Therefore, it is ideal to find an ensemble learning algorithm that 
can output diversified items~\cite{chen2018fast,li2020cascading,hiranandani2020cascading} and can directly optimize the online performance. 

Rank Aggregation (RA) is similar to ensemble learning. RA algorithm is a methodology to combine several permutations into a single permutation for a consensus rank, with applications in fields like web search, collaborative filtering, multi-agent planning and so on
. RA evolves from the social choice and voting theory, which focuses on combining individual preferences or interests to reach a collective decision. Since the RA problem is NP-hard even for only $4$ permutations~\cite{schalekamp2009rank} (i.e., it is hard to find the permutation with the lowest average distance even for only 4 input permutations\footnote{In this paper, we do not distinguish between sub-models, permutations, and voters.
}), it is often solved with heuristic and approximate algorithms. 

In our scenario, we replace the deep ensemble learning model with a contextual RA model, required to find proper weights of sub-models in different contexts. We name this novel framework the Learning-To-Ensemble (LTE) framework. In the framework, we need an algorithm that finds the set of weights for the contextual RA model to achieve the best online performance.  
Practically, because the inputs of RA models are scores produced by fine-tuned sub-models under a context, it is reasonable to assume that the best permutation is Pareto Optimal (PO). Under this assumption, we propose a new RA algorithm TournamentGreedy. TournamentGreedy has the best comprehensive performance amongst the known RA algorithms with quadratic time complexity. 

Finding the best weights for the sub-models can be regarded as a contextual Black-Box Optimization (BBO) problem. The classic BBO refers to finding a set of parameters to optimize an unknown function. As we need to include additional context information such as the gender of a user, it is non-trivial to apply classic BBO solutions in a straightforward manner. Unlike the classic BBO, contextual BBO allows modeling an additional group of uncontrollable input variables. These contexts are important: for example, male and female users do not share the same optimized weights. 
Classic Bayesian Optimization cannot handle the context unless it ignores the context, or separates the contextual BBO problems by context and solves several classic BBOs, which is difficult to optimize due to the data sparsity issue. 
We propose an Evaluator-Generator Optimization (EGO) framework for this contextual BBO problem. Inspired by Bayesian Optimization, EGO contains an Evaluator as the reward predictor and a Generator to generate the best weights under the supervision from the Evaluator along with an exploration bonus. Intuitively, EGO maps BBOs with different contexts into the same evaluation space. As the number of sub-models is small, EGO can generate proper weights to benefit the online system within an acceptable number of online interactions.

Our main contributions include:
\begin{itemize}
    \item To our knowledge, RA-EGO is the first practical LTE framework with a contextual RA model deployed on a large-scale E-commerce platform. The online A/B test shows it outperforms the industrial deep LTR solution in the ensemble task.
     \item As far as we know, the proposed EGO framework is the first to combine the classic Bayesian Optimization framework with deep learning to solve the contextual BBO problem.
    \item 
    We emphasize the importance of the expressive power of expected RA models and propose evaluation metrics on the expressive power. Under the assumption that the best permutation should be Pareto Optimal, we propose to examine algorithms by weak PO permutations, which can be efficiently computed to estimate the expressive power of algorithms.
    
    
   
\end{itemize}

\section{Related Works}\label{sec:rw}
In the early 20th century, Dwork et al.~\cite{dwork2001rank} use the RA technique to combat spam in Web Search. 
In his work, Borda's method~\cite{borda1784memoire}, Footrule approximation~\cite{bedo2014multivariate}, and different types of Markov Chain (MC) methods~\cite{freund2019rank} are experimented. 
Frans et al.~\cite{schalekamp2009rank} review RA methods and classify them into four categories, being positional, comparison sort, local search, and hybrid methods.
Ensemble learning has a different objective with RA. Ensemble learning models is generally a supervised learning model with inputs of sub-models, aim at predicting the conversion rate rather than finding the most consistent permutation as RA. In industry, the deep LTR model is a popular choice for ensemble learning~\cite{cao2007learning,burges2005learning,xia2008listwise,burges2010ranknet}, and the model ranks items according to their predicted scores when serves. 


BBO can offer feasible solutions for functions about which the analytical form is unknown. 
It has been widely studied and utilized in the field of hyper-parameter tuning, robotics planning, neural architecture search, material discovery and so on~\cite{meunier2020black}. Bayesian Optimization~\cite{pelikan1999boa} is a classic BBO algorithm~\cite{snoek2012practical}. In particular, Bayesian Optimization uses a Bayesian statistical model, such as Gaussian Process (GP) regression, to model the objective function and relies on an Acquisition Function (AF) to decide the next sample to interact with the system.
In many scenarios including ours, GP regression is not enough and needs to be replaced by kernel methods~\cite{duvenaud2013structure}, random forest~\cite{hutter2011sequential}, neural network~\cite{snoek2015scalable}, and others~\cite{tan2021cobbo}, which can be summarized as surrogate-based BBO methods~\cite{vu2017surrogate}.
Monte Carlo Tree Search (MCTS)~\cite{chaslot2008monte} is another popular algorithm for BBO used extensively in adversarial games and robotics planning.
Beomjoon et al. demonstrate the empirical effectiveness of the Voronoi partitioning algorithm for sampling, with the provision of the regret bound~\cite{kim2020monte}. To improve the sampling efficiency, Linnan et al. propose an MCTS based search algorithm that learns the action space for MCTS with recursive partitioning~\cite{wang2020learning}. 

\section{Preliminary}
In the standard \emph{RA} problem, there are $n$ \emph{permutations} $(p_1, p_2, ..., p_n)$ of $m$ candidates. The objective is to find the permutation with the lowest \emph{average weighted Kendall Tau Distance} (KTD) where the $i$-th permutation has weight $w_i$. Formally, we want to
\begin{equation}
    \arg \min_{p^* \in C} \frac{1}{n}\sum_{i=1}^{n}\left(w_i d_\tau(p^*, p_i)\right),
\end{equation}
where $C$ is the complete set of permutations with $m$ candidates. The function $d_\tau(p, q)$ is proportional to the number of inverse item pairs between permutations $p$ and $q$. Concretely, 
\begin{equation}
\begin{aligned}
    d_\tau(p, q) = \sum_{i=1}^m \sum_{j=i + 1}^m \frac{2 r\left(p^{-1}(i), p^{-1}(j), q^{-1}(i), q^{-1}(j)\right)} {m(m - 1)},
\end{aligned}
\end{equation}
\begin{equation}
    r(p_i, p_j, q_i, q_j) = \frac{1 - \text{sign}(p_i - p_j) \cdot { \text{sign}}(q_i - q_j)} {2},
\end{equation}
where $\text{sign}(x)$ denotes the sign function, and the inverse function of permutation, i.e., $p^{-1}(i)$, denotes the position of the $i$-th element in permutation $p$. Therefore, inverse pair function $r(p_i, p_j, q_i, q_j)$ equals $1$ when the order of elements $i$ and $j$ is different, i.e., $i$ is in the front of $j$ in one permutation but is in the behind of $j$ in the other permutation.

\subsection{Relation to Graph Theory}
A graph $G=(V,E)$ contains a set of vertices $V$ and a set of edges $E$. A \emph{tournament} is a standard notion in graph theory and can be explained as an undirected complete graph with an assignment of direction for each edge. A \emph{weighted tournament} is a tournament where each edge has a weight. In this paper, we denote a weighted edge as $(u, v, w)$ which points from $u$ to $v$ with the weight $w$. 
For easier presentation, we use $(u, v, \cdot)$ to denote an edge that points from $u$ to $v$ regardless of the weight. 
A \emph{directed acyclic graph} is a directed graph without cycle, implying that the \emph{topological sort} can be completed. The result of a topological sort is a permutation, in which $u$ is ranked before $v$ for each $(u, v)$ in the graph.
A \emph{minimal Feedback Arc Set (FAS) problem} requires us to remove edges to make the tournament a directed acyclic graph with minimal cost (i.e., the sum of weights on the removed edges). 
\begin{fact}
A standard rank aggregation problem can be reduced to a minimal FAS problem.
\end{fact}
This fact has been found in previous works~\cite{schalekamp2009rank,coleman2010ranking}. The key ingredient of the proof is to build a weighted tournament $(V, E)$, then the weights in the original rank aggregation problem are equivalent to the costs in the minimal FAS problem. We first build vertices $V = (1, 2, ..., m)$, where vertex $v$ represents the $v$-th candidate in the RA problem. Then we determine the edge direction of each pair of vertices $(v, u, \cdot)$ by the weighted frequency as follows
\begin{equation}\label{eq:edge}
\begin{aligned}
    &d(v, u) = \sum_{i=1}^n w_i \left(\text{sign}\left(p_i^{-1}(u) - p_i^{-1}(v)\right)\right), \\
    &(v, u, d(v, u)) \in E \iff d(v, u) > 0.
\end{aligned}
\end{equation}
If edge $(v, u, \cdot)$ exists, we set its weight as $d(v, u)$ and it represents the decreased cost when we let $v$ be in front of $u$ in the aggregation result. 
All weights are positive. If neither $(v, u,\cdot)$ nor $(u,v,\cdot)$ exists, we arbitrarily pick one with the weight of $0$. The permutation $\hat{p}$ obtained by the topological sort in the graph after removed the minimal FAS is the best permutation in the original RA problem. 


\subsection{Pareto Optimality}
\label{sec:prerec}
Given the input permutations $(p_1, p_2, ..., p_n)$ in an RA problem, we say a permutation $p$ is \emph{PO} if there is no permutation $q$ strictly better than $p$ on the metric of KTD. Concretely, 
\begin{equation}
    p \text{ is PO} \implies  \forall q \in C, \exists i, \text{ s.t. } d_\tau(p, p_i) \leq d_\tau(q, p_i).
    \label{eq:po}
\end{equation}
Practically, since the inputs are predictions from fine-tuned sub-models, it is reasonable to assume the permutation that yields the best online performance is in the set of PO permutations. Since the number of PO permutations is exponential, we weaken PO by observed permutations $C' \subset C$ in practice, then the Precision and Recall of an output set $S$ can be computed as
\begin{equation}
    P = \left\{\forall p \text{ satisfies Equation~\eqref{eq:po} for }q\in C'\text{ rather than }C\right\}
\end{equation}
\begin{equation}
    \text{Precision}(S) = \frac{\vert S \bigcap P\vert}{ \vert S \vert}, ~\text{Recall}(S) = \frac{\vert S \bigcap P\vert}{ \vert P \vert}
\end{equation}
Intuitively, Precision represents the probability that the output of an RA model is weak PO, and Recall represents the probability that a weak PO permutation can be outputted by an RA model with some weights. Both metrics indicate whether an RA model is likely to find the best permutation.

\subsection{Contextual BBO}
To optimize the online revenue, we regard weights selection as a BBO problem, where we optimize an unknown online revenue function $f(x)$. In an online environment, the best permutation heavily depends on the context, such as the gender and the age of a user. 
Therefore, we need to generalize BBO to include an uncontrollable input as the context. Formally, a contextual BBO contains an unknown conditional function $f(x|c)$ given a context distribution $C$ and we need to find a weight-generation policy $g$ such that $\arg\max_{g(\cdot)} \mathbb{E}_{c\sim C} \left[f\left(g(c)|c\right)\right]$.

It is hard to apply classic methods for contextual BBO unless we partition $f(x|c)$ into $\{f(x)\}_c$ and independently solve them as standard BBO tasks. However, it assumes the BBO with different contexts are Independent and Identically Distributed (i.i.d.) and drops the connections between data with different contexts. Furthermore, since the combinations of the context are exponential, it results in the data sparsity problem.

\begin{figure}[ht]
\centering
\includegraphics[width=0.42\textwidth]{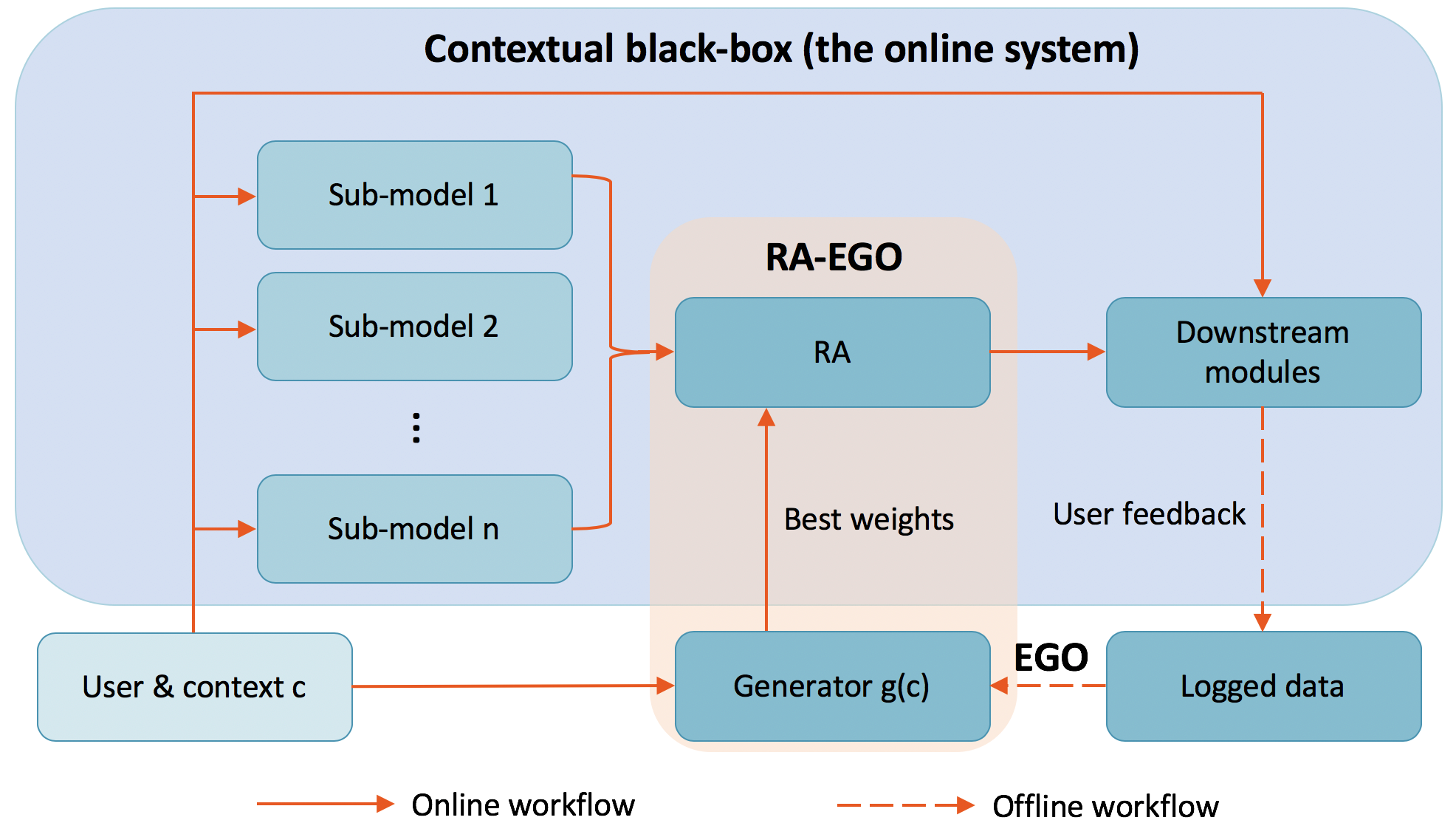}
\caption{The workflow of RA-EGO in the online system.}
\label{fig:onlinewf}
\end{figure}

\begin{figure*}[ht]
\centering
\subfigure[The illustration of process in TournamentGreedy. For the clear representation, the common factor $\sqrt{\frac{1}{\|V\|-1}}$ has been removed.]{
        \label{fig:tg}
        \includegraphics[scale=0.243]{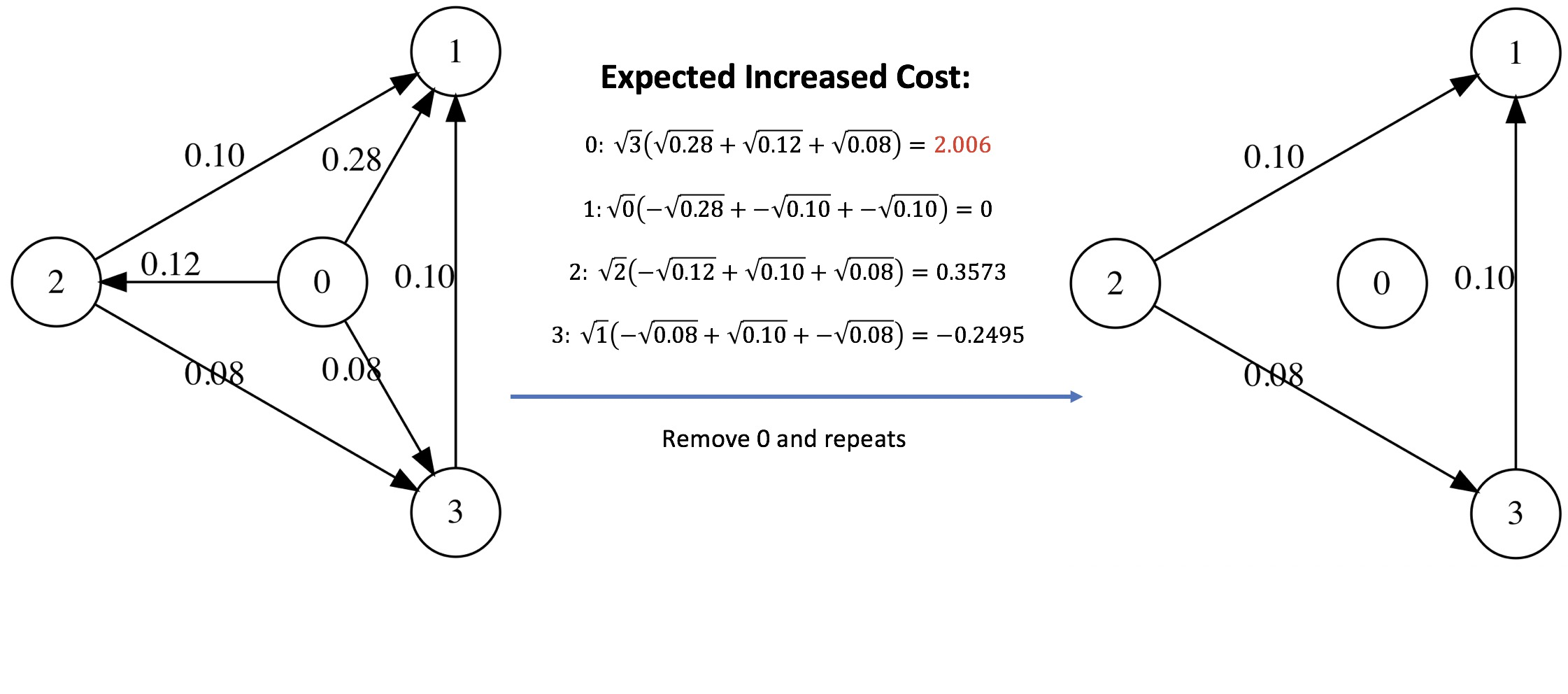}
}
\quad
\subfigure[The illustration of process in Evaluator-Generator Optimization.]{
        \label{fig:eg}
        \includegraphics[scale=0.23]{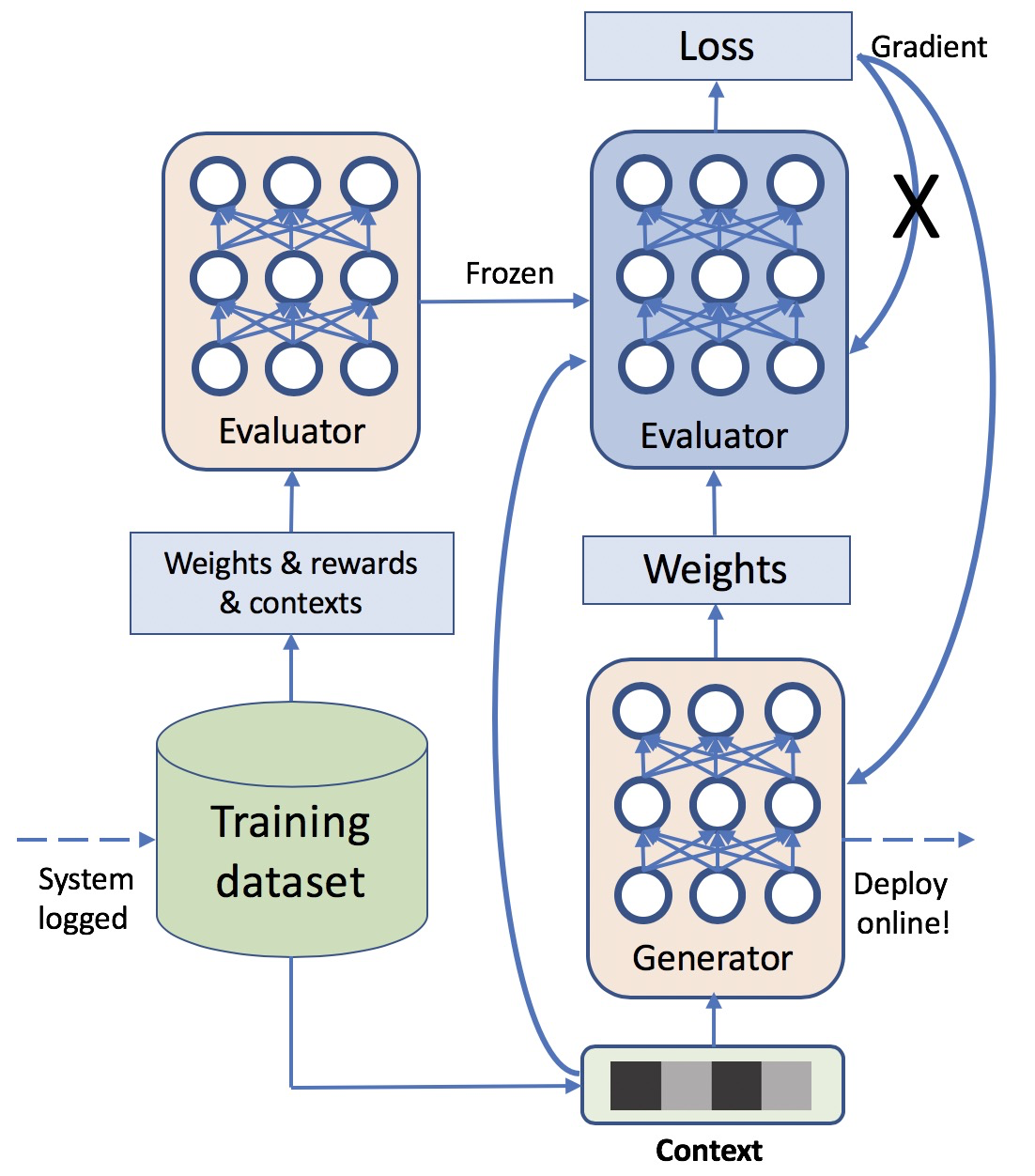}
}
\caption{The illustrations of TournamentGreedy and Evaluator-Generator Optimization.}
\end{figure*}

\section{Proposed Framework}
The high level idea of our framework is to use EGO, the proposed solution for contextual BBO, to solve the weights selection task. We describe the framework with pseudo code in Algorithm~\ref{alg:main} and its workflow in Figure~\ref{fig:onlinewf}. 

\begin{algorithm}
\caption{Learning-To-Ensemble via RA-EGO}
\label{alg:main}
\begin{algorithmic}
\STATE {\textbf{Notation: } Online feedback $f_{c, \text{R}}$ maps weights set $W = (w_1, w_2, ..., w_n)$ to rewards $r$ with the given context $c$ and the rank aggregator $R$, weight generation policy $g$ maps the context $c$ to an optimized set of weights $W$}
\STATE {} 
\STATE {// Cold start}
\STATE {Dataset $\mathbb{D}\gets \emptyset$}
\STATE {$\text{R} \gets \text{TournamentGreedy} $ // Refer Section~\ref{sec:ra}}
\STATE {Generate $k$ weights sets $\{W_1, W_2, ..., W_k\}$ by experts, weight generation policy $g$ randomly picks one}
\STATE {$\mathbb{D}\gets \mathbb{D}\cup\left\{g(c), f_{c, \text{R}}\left(g(c)\right)\right\}_c$ // Expensive online operation}
\STATE {// Warm start} 
\STATE {Weight generation policy $g \gets \text{EGO}(\mathbb{D})$  // Refer Section~\ref{sec:ego}}
\STATE {$\mathbb{D}\gets \mathbb{D}\cup\left\{g(c), f_{c, \text{R}}\left(g(c)\right)\right\}_c$ // Expensive online operation}
\end{algorithmic}
\end{algorithm}
\vspace{-1.5em}

\subsection{Proposed RA Model}
\label{sec:ra}
One important difference between the classic RA problem and ours is that 
the classic objective to minimize the average weighted KTD cannot promise the best online revenue. 
In our framework, we emphasize the importance of \emph{the expressive power} of an RA model. With proper weights, the designed RA model can possibly express the permutation that could generate the highest revenue online.

Therefore, besides KTD, we also study two properties of RA models: \emph{Efficiency} (how likely the generated permutations are to be the best), and \emph{Coverage} (how many of the permutations, that potentially will optimize the online revenue, can be generated by an RA model). 
We introduce the measurement \emph{Precision} and \emph{Recall} to evaluate the above two properties of  RA models. Our TournamentGreedy has the best average weighted KTD among all the considered RA models, and has a high performance in Precision and Recall. The performances can be found in the experimental section.

The high level idea of TournamentGreedy is to greedily decide the next element by minimizing \emph{the expected increased cost}. We build the tournament $G=(V,E)$ following Equation~\eqref{eq:edge}. Pessimistically, we compute the increased cost of choosing $i$ as $\sum_{(v,i,w)\in E} w - \sum_{(i,v,w)\in E} w$, which is the difference of summation of weights that places $i$ in the next position or the last position. The expected increased cost $c(i)$ of element $i$ is a smooth combination of the probability that $v$ is chosen where $(v, i)$ exists and the above cost
\begin{equation}\label{eq:ic}
    c(i)=\sqrt{\frac{\sum_{(v,i,w)\in E} \mathbb{I}(w > 0)}{\|V\| - 1}} \cdot \left(\sum_{(v,i,w)\in E} \sqrt{w} - \sum_{(i,v,w)\in E} \sqrt{w}\right),
\end{equation}
where $\mathbb{I}(w>0)=1$ if $w>0$. We provide an example on how TournamentGreedy works in Figure~\ref{fig:tg} and the corresponding pseudo code is listed in Algorithm~\ref{alg:tg}. Note that $c(i)$ can be updated in linear time without explicit changing the graph $(V, E)$. Therefore, the time complexity of the proposed RA model is  $O(nm + m^2)$, which includes $O(nm)$ for graph building and $O(m^2)$ for greedy decisions.


\begin{algorithm}
\caption{TournamentGreedy}
\label{alg:tg}
\begin{algorithmic}
\STATE {\textbf{Input: } $n$ permutations and their weights $\{(p_i, w_i)\}_{i=1}^{n}$, the number of candidates $m$}
\STATE {\textbf{Output: } a permutation $\hat{p}$}

\STATE{}
\STATE {$V \gets [m]$, $E \gets \emptyset$} 
\FOR {$i, j \in [m], i\neq j$} 
\STATE {$\text{weightSum} \gets 0$}
\FOR {$k \in [n], k\neq j$} 
\IF {$i$ is in the front of $j$ in $p_k$}
\STATE {$\text{weightSum} \gets \text{weightSum} - w_k$ }
\ELSE
\STATE {$\text{weightSum} \gets \text{weightSum} + w_k$ }
\ENDIF
\ENDFOR
\STATE {$E \gets E \cup (i, j, \text{weightSum})$}
\ENDFOR
\STATE{Initialize permutation $\hat{p}$ as an empty permutation}
\FOR {$t \in [m]$} 
        \STATE {Put $i$ to the end of $\hat{p}$ with the maximal $c(i)$ in Equation~\eqref{eq:ic}}
        \STATE {$V \gets V \setminus \{i\}$ , $E \gets E \setminus (\{(i, \cdot, \cdot)\}\cup\{(\cdot, i, \cdot)\})$ }
\ENDFOR
\RETURN{$\hat{p}$}
\end{algorithmic}
\end{algorithm}
\vspace{-0.5em}

\begin{figure*}[ht]
\label{fig:egaf}
\centering
\subfigure[The ground-truth of reward that TournamentGreedy produces with weights $(x, y, z)$.]{
        \label{fig:egaf1}
        \includegraphics[width=0.2\textwidth]{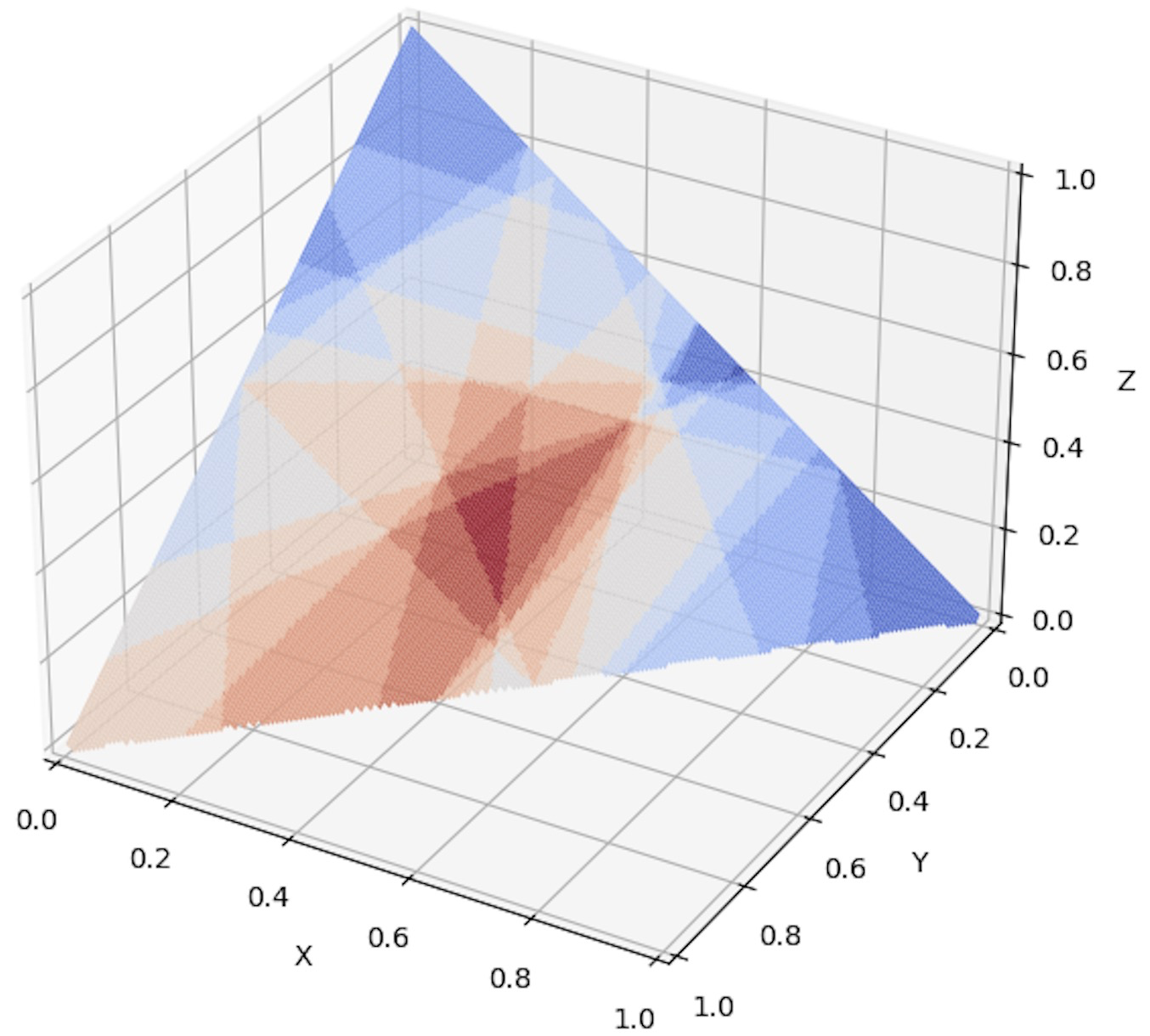}
}
\quad
\subfigure[Reward prediction with pure exploitation strategy after 20 rounds. The latest weights is marked in green.]{
        \label{fig:egaf2}
        \includegraphics[width=0.2\textwidth]{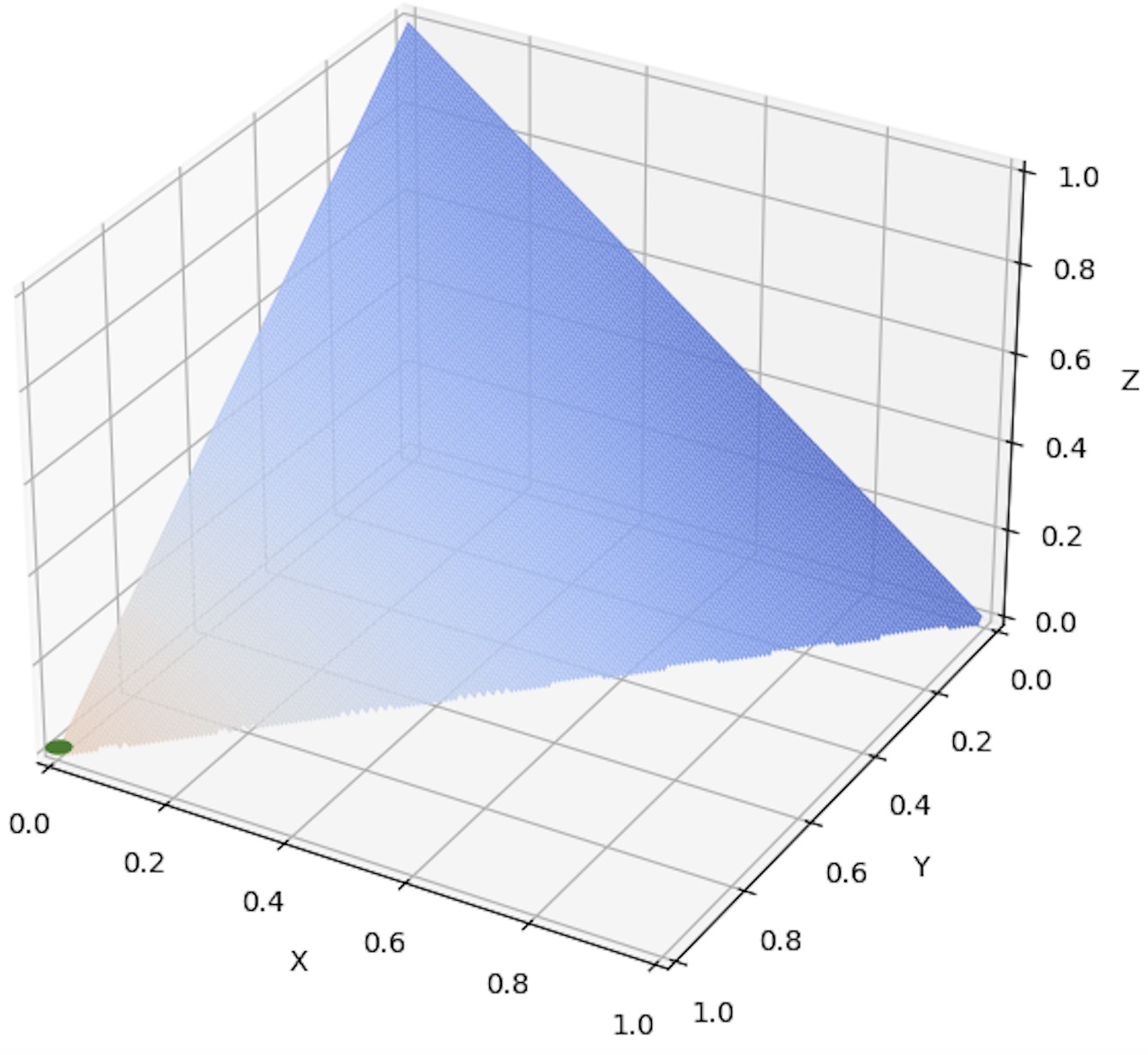}
}
\quad
\subfigure[Reward prediction with exploration bonuses after 20 rounds. ]{
        \label{fig:egaf3}
        \includegraphics[width=0.2\textwidth]{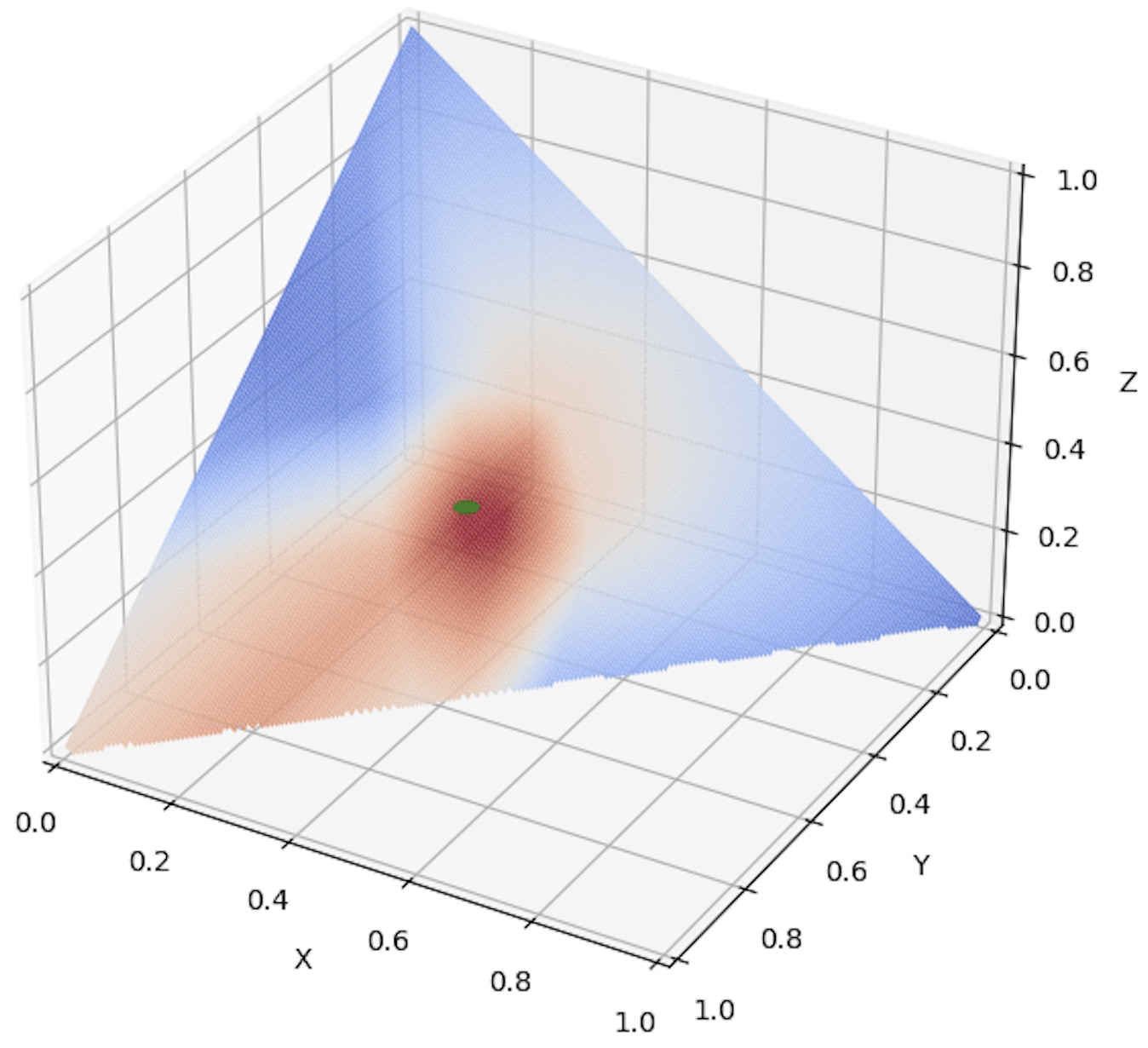}
} 
\quad
\subfigure[The distribution of exploration bonus after 20 rounds.]{
        \label{fig:egaf4}
        \includegraphics[width=0.2\textwidth]{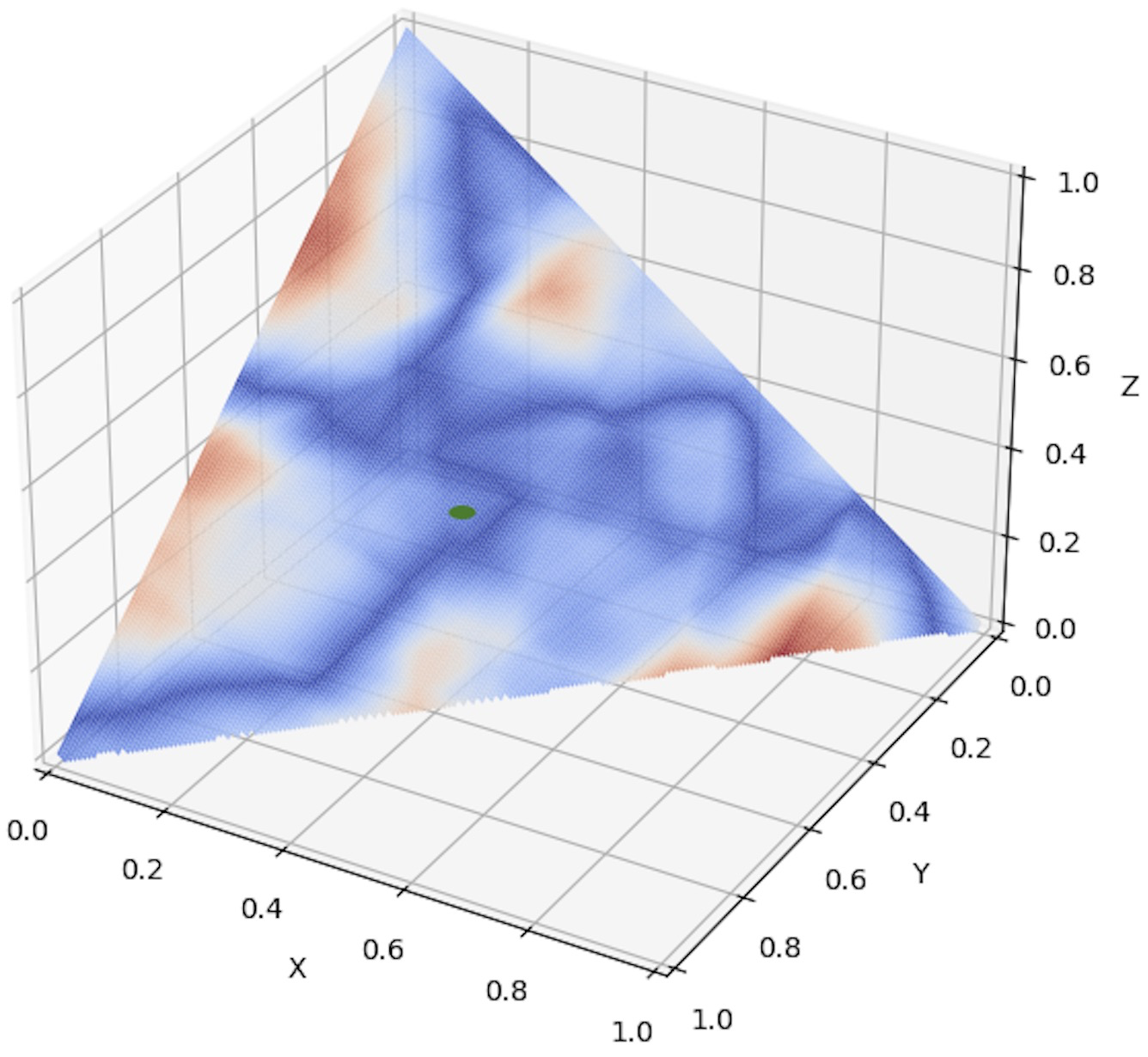}
}
\caption{The importance of the exploration bonus in EGO. Hotter area implies the greater reward.}
\label{fig:af}
\end{figure*}

\subsection{Weights Optimization as Contextual BBO}
\label{sec:ego}
Without context, the weight optimization task can be regarded as a standard BBO problem. We combine the framework of Bayesian Optimization and neural networks for the generalization ability. Similar to the Bayesian Optimization approaches, ours has a \emph{surrogate model} to estimate the reward and an \emph{Acquisition Function (AF)} to search the solution space. The general choice of the surrogate model is Gaussian Process (GP), and the general choice for AF is the Upper Confidence Bound (UCB) function. However, it is difficult for GP to handle contexts. A possible solution is to apply classic Bayesian Optimization by partitioning the original problem into sub-problems according to their contexts. However, as there are many permutations of the context variables, the number of sub-problems is large. Thus each sub-problem may have few data points and difficult to optimize. Hence we propose the \emph{Evaluator}, a deep surrogate model which inputs a context vector and the weights for the RA model. 
One difficulty with adopting the deep surrogate model is optimal solution searching. A straightforward idea is to perform multiple gradient descents to get the optimal weights, but it is time-consuming when deployed online.
In EGO, we prepare another deep model, which we call the \emph{Generator}, to optimize the feedback of Evaluator and quickly infer weights of sub-models online. The complete process can be found in Figure~\ref{fig:eg} and the pseudo code can be found in Algorithm~\ref{alg:ego}.

\subsubsection{The Evaluator}
The inputs to the evaluator is the context and weights of sub-models. 
We use a simple Multi-Layer Perceptron (MLP) $M$ with layer sizes $64$ and $32$ to classify whether the input weights will have a better performance compared to current online weights.
With a well-trained Evaluator, the model weights can be correctly predicted with history data. Yet there is a need to explore unseen permutations as the corresponding parameters with the new samples can possibly result in better performance than the current prediction.  
To give exploration bonuses on the unreliable space of the Evaluator, we borrow the idea from Random Network Distillation (RND)~\cite{burda2018exploration}. We additionally set up an auxiliary regression task $R$ with an MLP model $M_R$, whose inputs are the same as the Evaluator but outputs are vectors and its oracle is a randomly weighted MLP $M_R^*$. Intuitively, this auxiliary task characterizes confidence on the prediction of the Evaluator: given parameters $x$, if the prediction $M_R(x)$ is far different with the oracle $M_R^*(x)$, the prediction of the Evaluator may have a similar difference to the true reward.
Let $l(x) \in [-1,1]$ denote the label about the conversion rate gap in comparison to the baseline, $\mathbb{I}\left(l(x) > 0)\right)$ denotes the corresponding $0$-$1$ label, and $s\left(l(x)\right)$ denotes the number of samples with parameters $x$.
Thus, the objective function of the Evaluator can be represented as
\begin{equation}
\begin{aligned}
    \arg\min_{M, M_R} \sum_{x} \text{WCE}\left(M(x), l(x)\right) + \text{MSE}\left(M_R(x), M_R^*(x)\right), \\
    \text{WCE}\left(M(x), l(x)\right) = \text{CE}\left(M(x), \mathbb{I}(l(x) > 0)\right) \cdot s\left(l(x)\right) \cdot |l(x)|,
\end{aligned}
\end{equation}
where $\text{CE}$ and $\text{MSE}$ denote the Cross-Entropy (CE) loss and the Mean-Squared-Error (MSE) loss, respectively.

\begin{algorithm}
\caption{The Evaluator-Generator Optimization}
\label{alg:ego}
\begin{algorithmic}
\STATE {\textbf{Input: } Dataset $\mathbb{D} = \{(c_i, w_i, r_i)\}$ contains contexts, weights of sub-models and rewards.}
\STATE {\textbf{Output: } The generation policy $g(c)$ which aims at producing optimized weights of sub-models on a given context $c$.}

\STATE{}
\STATE{Initialize the Evaluator $M(c, p)\rightarrow r$ and train it by supervised learning with dataset $\mathbb{D}$}
\STATE{Initialize the Evaluator $M_R(c, p),~M_R^*(c, p)\rightarrow \mathbf{v}$ with random weights, where virtual target $\mathbf{v}$ is a vector}
\STATE{Train $M_R$ by supervised learning with modified dataset $\mathbb{D'} = \{(c_i, w_i, M_R^*(c_i, w_i))\}$}
\STATE{Train the parameters $\theta_g$ of the Generator $g(c) \rightarrow p$ by supervised learning with dataset $\mathbb{D}$ and loss $L(\theta_g)$}

\RETURN{Generator $g$}
\end{algorithmic}
\end{algorithm}
\vspace{-0.5em}

\begin{figure*}[ht]
\centering
\includegraphics[width=0.9\textwidth]{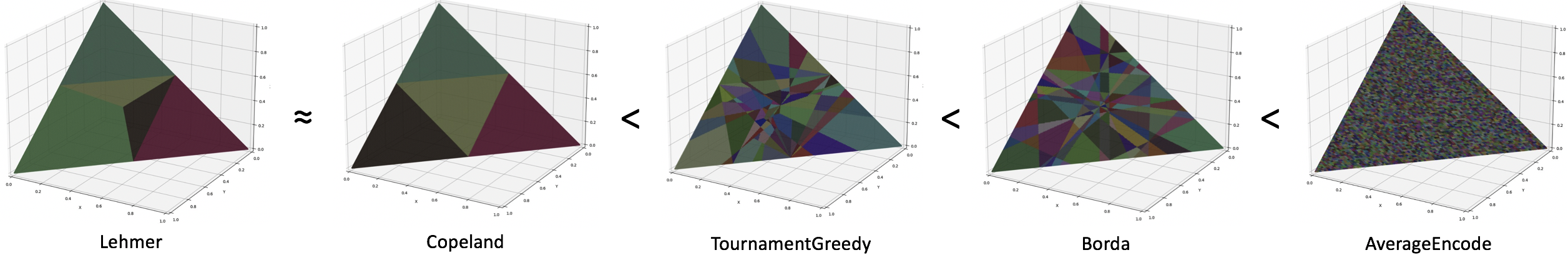}
\caption{Virtualization on diversity in a toy example with $3$ permutations.}
\label{fig:div}
\end{figure*}

\subsubsection{The Generator}
After the Evaluator $M$ and the auxiliary models $M_R$ and $M_R^*$ are trained, the Generator $g$ aims at exploring parameters, i.e., the weights of sub-models given a context $c$, which can maximize the following reward
\begin{equation}\label{eq:gen}
    \arg\max_g \sum_c M\left(g(c)\right) + \alpha \left\|M_R\left(g(c)\right) - M_R^*\left(g(c)\right)\right\|^2.
\end{equation}
The above formula is similar to UCB in Multi-Arm Bandit (MAB) applications: the first term is the average reward and the second term is the confidence bound (i.e. exploration bonus). In our scenario, the Generator $g$ is an MLP and can be parameterized by $\theta_g$.

As $M$, $M_R$ and $M_R^*$ are differentiable, let
\begin{equation}
    L(\theta_g) = -\sum_c M\left(g(c)\right) - \alpha \left\|M_R\left(g(c)\right) - M_R^*\left(g(c)\right)\right\|^2,
\end{equation}
and Back-Propagation (BP) is allowed to find the best $\theta_g^*$, with stopping updates of gradients to parameters within $M$, $M_R$ and $M_R^*$. The BP process is gradient descent of vector $\theta_g$ in the value space of $L(\theta_g)$. We summarize the above process in Algorithm~\ref{alg:ego}.

To demonstrate the effectiveness of the exploration bonus brought by the auxiliary task, we set up a toy example and study the accuracy of the reward predictions without involving $M_R$ and $M_R^*$. Unlike the reward function of common black-box optimizations, the values of the reward function in our scenario are formed by several connected regions, with each region sharing the same reward value as Figure~\ref{fig:egaf1} shows. Therefore, a pure exploitation strategy that always chooses the permutation with the best response from the reward predictor $M$ may easily fall into the local optima.
In this toy example, we have $3$ target permutations and need to find the best weights $(x,y,z)$ for TournamentGreedy to optimize a designed reward function, as described in Figure~\ref{fig:egaf1}. Initially, we give the ground-truth rewards at weights $(1, 0, 0)$, $(0, 1, 0)$, and $(0, 0, 1)$ and each model has $20$ rounds to interact with the reward function.
As Figure~\ref{fig:egaf2} shows, the pure exploitation strategy sticks at the local optima and always outputs weight $(0, 1, 0)$, which is the best among three initial combinations of weights. On the other hand, the reward predictor can roughly reproduce the ground-truth in a short time with the exploration bonus as Figure~\ref{fig:egaf3} shows.

\section{Experiments}
We design experiments to examine the following questions:
\begin{itemize}
    \item \textbf{RQ1:} Does TournamentGreedy outperform the classic ones for random voters?
    \item \textbf{RQ2:} Does TournamentGreedy outperform the classic algorithms with preferences from real-world users?
    \item \textbf{RQ3:} Does our proposed ensemble framework RA-EGO outperform the industrial deep ensemble learning model?
\end{itemize}

\subsection{Offline Experiment}
We examine our proposed RA model in two kinds of offline environments: one is the randomly generated dataset with different scales of voters and candidates, and the other is a group of real-world datasets extracted from thousands of customer ratings.

\subsubsection{Baselines}
Since a real-world system requires a short response time, we only consider efficient RA models and the chosen baselines include $4$ classic RA models with computational complexity lower than $\mathcal{O}(nm+m^2)$, which are listed as follows.
\begin{itemize}
    \item \textbf{Dictator}. It always picks the highest weighted permutation.
    \item \textbf{Borda's method}. It computes the average position of each element as the Borda score and then ranks elements according to their Borda scores in ascending order~\cite{schalekamp2009rank}.
    \item \textbf{Lehmer Code method}. This method transforms the input permutations into their Lehmer code domain and selects the weighted mode of items to produce the final result~\cite{li2017efficient}.
    \item \textbf{Copeland's method}. Firstly, define a relation $\prec$ on elements, where $i \prec j$ if more than half of sub-models agree on that $i$ should be ranked before $j$. Secondly, define the Copeland score of an element $i$ to be $\left| \left\{ y: x \prec y \right\} \right|$. Then the Copeland's method ranks elements according to their Copeland scores in descending order~\cite{schalekamp2009rank}.
\end{itemize}

\subsubsection{Virtualization on Diversity}
We define Diversity as the number of distinct permutations that an RA model can generate. We set up a toy example with $3$ permutations and show the Diversity of different methods in Figure~\ref{fig:div}. The plotted plane is $x+y+z=1$ for permutation weights $x,y,z>0$. The different colors in this figure represent the different output permutations. The AverageEncode uses Cantor expansion to encode permutations by integers, and then decode the average round integer back to the result permutation. It can be observed that TournamentGreedy and Borda's method can produce significantly more permutations with proper weights than Lehmer Code and Copeland's methods. AverageEncode produces diversified permutations and its Diversity metric can serve as the upper bound, yet it does not optimize the online revenue because its outputs are almost random.

\begin{table*}[ht]
\small
\resizebox{.96\textwidth}{!}{
\begin{tabular}{ll|ll|ll|ll|ll|ll|ll}
\toprule \hline
 &  & \multicolumn{6}{c|}{\textbf{Uniform Weight Voters}}   & \multicolumn{6}{c}{\textbf{Random Weight Voters}}  \\
 &  & \multicolumn{2}{c}{\textbf{8 candidates}} & \multicolumn{2}{c}{\textbf{20 candidates}} &\multicolumn{2}{c|}{\textbf{50 candidates}} & \multicolumn{2}{c}{\textbf{8 candidates}}& \multicolumn{2}{c}{\textbf{20 candidates}} & \multicolumn{2}{c}{\textbf{50 candidates}} \\\hline
 &  & Efficiency & Fairness & Efficiency & Fairness & Efficiency & Fairness & Efficiency & Fairness & Efficiency & Fairness & Efficiency & Fairness\\ \hline
\multirow{5}{*}{\textbf{3 voters}}  & Dictator & 0.333139 & 0.166633 & 0.333536 & 0.166798 & 0.333159 & 0.166619 & 0.333686 & 0.146228 & 0.333834 & 0.184139 & 0.129505 & 0.131653 \\
 & Copeland & 0.278733 & 0.093029 & 0.290340 & 0.096859 & 0.295322 & 0.098570 & 0.306220 & 0.134580 & 0.311620 & 0.172000 & 0.117496 & 0.124113 \\
 & Lehmer & 0.351800 & 0.117316 & 0.381537 & 0.127364 & 0.392668 & 0.130960 & 0.330464 & 0.144903 & 0.335883 & 0.185227 & 0.127880 & 0.132648 \\
 & Borda & 0.290815 & 0.097030 & 0.298397 & 0.099528 & 0.300922 & 0.100390 & 0.310177 & 0.135934 & 0.313572 & 0.173083 & 0.117284 & 0.124582 \\
 & \textbf{Proposed} & \textbf{0.273848} & \textbf{0.091303} & \textbf{0.287520} & \textbf{0.095944} & \textbf{0.294981} & \textbf{0.098432} & \textbf{0.303375} & \textbf{0.133296} & \textbf{0.309019} & \textbf{0.170583} & \textbf{0.116853} & \textbf{0.123413} \\\hline
\multirow{5}{*}{\textbf{10 voters}} & Dictator & 0.450368 & 0.050148 & 0.449943 & 0.050045 & 0.450097 & 0.050038 & 0.450229 & 0.121667 & 0.449989 & 0.094516 & 0.450293 & 0.129505 \\ 
 & Copeland & 0.390515 & 0.039157 & 0.393146 & 0.039373 & 0.394614 & 0.039496 & 0.403744 & 0.109014 & 0.407007 & 0.085557 & 0.408583 & 0.117496 \\
 & Lehmer & 0.420247 & 0.042083 & 0.434999 & 0.043570 & 0.449699 & 0.045009 & 0.426836 & 0.115400 & 0.437725 & 0.092045 & 0.444994 & 0.127880 \\
 & Borda & 0.389644 & 0.039050 & 0.392940 & 0.039364 & 0.394712 & 0.039507 & 0.403352 & 0.108845 & 0.406409 & 0.085444 & 0.407941 & 0.117284 \\
 & \textbf{Proposed} & \textbf{0.383025} & \textbf{0.038417} & \textbf{0.388549} & \textbf{0.038919} & \textbf{0.392431} & \textbf{0.039282} & \textbf{0.397740} & \textbf{0.107396} & \textbf{0.403145} & \textbf{0.084698} & \textbf{0.406444} & \textbf{0.116853} \\\hline
\multirow{5}{*}{\textbf{30 voters}} & Dictator & 0.483299 & 0.016724 & 0.483403 & 0.016692 & 0.483341 & 0.016679 & 0.483307 & 0.028179 & 0.483308 & 0.038145 & 0.483290 & 0.026520 \\ 
 & Copeland & 0.436958 & 0.014608 & 0.438938 & 0.014655 & 0.439702 & 0.014670 & 0.444987 & 0.025931 & 0.446949 & 0.035302 & 0.447776 & 0.024564 \\
 & Lehmer & 0.455614 & 0.015235 & 0.464364 & 0.015528 & 0.471600 & 0.015733 & 0.460531 & 0.026805 & 0.467795 & 0.036967 & 0.472719 & 0.025941 \\
 & Borda & 0.436693 & 0.014603 & 0.438808 & 0.014650 & 0.439697 & 0.014668 & 0.444960 & 0.025917 & 0.446811 & 0.035295 & 0.447697 & 0.024561 \\
 & \textbf{Proposed} & \textbf{0.432597} & \textbf{0.014479} & \textbf{0.436291} & \textbf{0.014566} & \textbf{0.438427} & \textbf{0.014627} & \textbf{0.441501} & \textbf{0.025713} & \textbf{0.444803} & \textbf{0.035133} & \textbf{0.446710} & \textbf{0.024508} \\ \hline
\bottomrule
\end{tabular}}
\caption{Efficiency and Fairness comparisons of the proposed and baseline algorithms with uniform and random weight voters.}\label{tab:cmp_1}
\vspace{-1em}
\end{table*}

\begin{table*}[ht]
\small
\resizebox{.78\textwidth}{!}{
\begin{tabular}{ll|lll|lll|lll}
\toprule \hline
 &  & \multicolumn{3}{c|}{\textbf{8 candidates}} & \multicolumn{3}{c|}{\textbf{20 candidates}} &\multicolumn{3}{c}{\textbf{50 candidates}} \\\hline
 &  & Diversity & Precision & Recall & Diversity & Precision & Recall & Diversity & Precision & Recall \\ \hline
\multirow{5}{*}{\textbf{3 voters}} & Dictator & 0.032835 & \textbf{1.000000} & 0.074078 & 0.000569 & \textbf{1.000000} & 0.005141 & 0.000019 & \textbf{1.000000} & 0.000297 \\
&Copeland & 0.051164 & 0.871000 & 0.105351 & 0.000948 & 0.624000 & 0.005364 & 0.000031 & 0.600000 & 0.000297 \\
&Lehmer & 0.054122 & 0.669143 & 0.084747 & 0.001170 & 0.504667 & 0.005258 & 0.000039 & 0.493048 & 0.000300 \\
&Borda & \textbf{0.709906} & 0.386323 & 0.660930 & \textbf{0.682931} & 0.046277 & 0.285271 & \textbf{0.721840} & 0.007076 & 0.080663 \\
&\textbf{Proposed} & 0.519149 & 0.514662 & \textbf{0.662682} & 0.319168 & 0.110721 & \textbf{0.314686} & 0.278190 & 0.019098 & \textbf{0.083801} \\\hline
\bottomrule
\end{tabular}}
\caption{Diversity, Precision, and Recall comparisons of the proposed and baseline algorithms with the uniform weight voters.}\label{tab:cmp_2}
\vspace{-1em}
\end{table*}

\subsubsection{Measurements}
In this subsection, we consider the following measurements for our proposed and baseline RA models. As it is impossible to enumerate all PO permutations, we use weak PO permutations to examine the expressing power of RA models.
\begin{itemize}
    \item \textbf{Efficiency on KTD.} Efficiency is a common economic notion that describes how a whole society benefits. Averaged KTD is a standard metric that is commonly used in examinations of RA models. It can represent Efficiency in our scenario and can be formalized as $\text{Eff}(p_\text{r}) = \frac{1}{n} \sum_{i=1}^{n} w_i d_\tau(p_\text{r}, p_i)$,
    where $p_\text{r}$ denotes the output of the RA model with input permutations $\{p_1, \dots, p_n\}$ and their weights $\{w_1, \dots, w_n\}$.
    As $\text{Eff}(p_\text{r})$ is measured by KTD, the lower $\text{Eff}(p_\text{r})$ indicates that the better global similarity is achieved.
    \item \textbf{Fairness on KTD.} The relationship between fairness and efficiency is usually adversarial and we need to obtain a trade-off between them. In our setting, we use the largest KTD to represent fairness corresponding to an RA model, i.e., $\text{Fair}(p_\text{r}) = \max_{i=1,\dots,n}\{w_i d_\tau(p_\text{r}, p_i)\}$.
    The lower $\text{Fair}(p_\text{r})$ implies the better order consistency to the result permutation is achieved for the poorest permutation.
    \item \textbf{Diversity.} In this experiment, we count the number of distinct permutations that an RA model produces with one million random weights as the Diversity.
    \item \textbf{Precision on weak PO solutions. } 
    Precision measures how precisely an RA model can produce weak PO permutations. A larger Precision value implies the output of the RA model is more likely to be a weak PO permutation. The formula of Precision and Recall can be found in Section~\ref{sec:prerec}.
    \item \textbf{Recall on weak PO solutions. } Recall value is the ratio that weak PO permutations can be produced by an RA model. A larger Recall value implies the RA model can eventually find the optimized permutation with sufficient explorations.
\end{itemize}


\subsubsection{Random Benchmark}
A random benchmark is sufficient for offline RA models evaluation and its results can be easily reproduced. In each round, the environment randomly generates $n$ different permutations $\{p_i\}$ among $m$ candidates and generate a randomly weight $w_i\in[0,1]$ for each permutation, where $w_i$ will be normalized by $w_i / \sum w_i$. After that, an RA model needs to output a permutation with inputs $\{(p_i, w_i)\}$. 
In our work, we examine the performance of baseline methods and the proposed one in different settings with various scales of votes and candidates. Concretely, we examine methods in 9 independent settings of the random benchmark with 8, 20, 50 candidates and 3, 10, 30 voters. 

\textbf{Efficiency and Fairness.} The Efficiency and Fairness comparisons are measured in both uniform and random weight voters setting. The uniform weight voters are a setting in the classic baseline methods and the random weight voters setting reflects special considerations for E-commerce scenarios. The experiment results on Efficiency and Fairness are listed in Table~\ref{tab:cmp_1}, with average $50000$ samples of random permutations. Our TournamentGreedy outperforms all the baselines on measurements Efficiency and Fairness, under uniform and random weight voters settings.

\textbf{Diversity, Precision, and Recall.} The Diversity, Precision, and Recall comparisons are displayed in Table~\ref{tab:cmp_2}, in which $1$ million samples are randomly generated to obtain the weak PO set. 
To better relieve the bias brought by replacing PO with weak PO, we only examine the performance of RA models under the setting of 8 voters.
As the Dictator method always selects the permutation with the largest weight, it has a Precision of up to $100\%$. However, it is far less diverse than Borda's method and TournamentGreedy. Although the proposed TournamentGreedy has lower Diversity than Borda's method, its Precision and Recall are relatively higher. Overall, among all the considered RA models with complexity lower than $\mathcal{O}(nm+m^2)$, the proposed TournamentGreedy method achieves the best trade-off among Diversity, Precision, and Recall.

\subsubsection{Real-World Dataset}
As the behavior pattern of real-world voters may be different from the random permutations, we examine our method in two real-world datasets listed in~\cite{li2017efficient}, the Jester dataset and Sushi dataset. Jester dataset contains $6.5$ million anonymous ratings of $100$ jokes by the Jester Joke Recommender System users and we take out totally $14116$ ratings that cover all of the $100$ jokes. Similarly, Sushi dataset includes $5000$ orders of $10$ kinds of top rated sushi. Following the examination methodology in ~\cite{li2017efficient}, we examine TournamentGreedy and the classic RA 
models by randomly choosing $50$, $200$, $1000$, $5000$, $10000$ permutations from each dataset. The Efficiency metric is calculated by averaging $50$ independently experiment results. From comparisons listed in Table~\ref{tab:cmp_3}, we find the proposed TournamentGreedy method has the best Efficiency performance in most cases. But as the number of voters increases, the gap between the baselines and the proposed one diminishes and Copeland's method obtains better Efficiency at $5000$ and $10000$ voters on Jester Dataset.

\begin{table*}[ht]
\small
\resizebox{.99\textwidth}{!}{
\begin{tabular}{l|lllll|lllll}
\toprule \hline
 & \multicolumn{5}{c|}{\textbf{Jester Dataset}} & \multicolumn{5}{c}{\textbf{Sushi Dataset}} \\\hline
 \textbf{Voters} & \textbf{50} & \textbf{200} & \textbf{1000} & \textbf{5000} & \textbf{10000}  & \textbf{50} & \textbf{200} & \textbf{1000} & \textbf{5000} & \textbf{10000} \\ \hline
Dictator & 0.455278 & 0.461552 & 0.458725 & 0.432150 & 0.440028 & 0.475591  & 0.479541  & 0.485994  & 0.483410  & 0.478364 \\
Copeland & 0.378942 & 0.385276 & 0.387709 & \textbf{0.386687} & \textbf{0.387343} & 0.413360  & 0.421786  & 0.425510  & 0.426090  & 0.426046 \\
Lehmer & 0.418019 & 0.407635 & 0.405217 & 0.405607 & 0.404071 & 0.424053  & 0.425423  & 0.427978  & 0.428867  & 0.428902 \\
Borda & 0.379292 & 0.385460 & 0.387830 & 0.386809 & 0.387454 & 0.414418  & 0.422401  & 0.425739  & 0.426169  & 0.426114 \\
\textbf{Proposed} & \textbf{0.378818} & \textbf{0.385214} & \textbf{0.387704} & 0.386690 & 0.387350 & \textbf{0.411867}  & \textbf{0.421481}  & \textbf{0.425462}  & \textbf{0.426064}  & \textbf{0.426045} \\\hline
\bottomrule
\end{tabular}}
\caption{Efficiency comparisons of algorithms on two real-world datasets under the uniform weight voters setting.}\label{tab:cmp_3}
\vspace{-1em}
\end{table*}

\subsection{Online Experiment}
Since a surrogate objective always has a gap with the online revenue,
online A/B test is the only golden criterion for our comparisons. We set up online A/B tests on the AliExpress Search System to examine our proposed RA-EGO. 
We prepare $4$ methods in the online A/B test, namely Deep LTR, RA-RE, RA-BE, and RA-EGO. Deep LTR  follows a recent work HMOE~\cite{li2020improving} as a classic deep LTR model for the ensemble task with inputs from more than 40 sub-models. It has been implemented to update its weights in real-time on our online platform. 
RA-RE is the weight generation policy for cold-start in the Algorithm~\ref{alg:main}, and it will choose a random combination of weights from a set designed by human experts. The designed set of weights contains $8$ different combinations of weights on several main metrics including the click-through rate, the conversion rate, relevance scores between user and items, sales of items, and so on.
RA-BE relies on data collected by the RA-RE method, and always selects the combination of weights with the highest accumulated online conversion rates in real-time.
RA-EGO is our proposed contextual method utilizing EGO for sub-model weights generation and TournamentGreedy for item rank aggregation. 

Furthermore, in the E-Commerce scenario, users have potential behavior patterns on each position of a given rank, well-known as \emph{Position Bias}~\cite{joachims2017unbiased,guo2019pal}. Sub-models could have differing predictive power at different positions. For example, some sub-models have better predictive power at top positions and others are better at tail positions.
To better model the positions, we modify the weights of $i$-th sub-model on the $k$-th position as $\omega_i(k) = w_i \gamma_i^k$,
in which the $i$-th rank has the original weight $w_i$, and the adjustable decay factor $\gamma_i$ denotes the positional decay factor of $i$-th sub-model. Therefore, less $\gamma_i$ implies the model influences less on the decision of the latter items. Under this setting, the Generator needs to produce a number of weights equal to twice the number of sub-models.


The above $4$ methods serve a non-overlapping random portion of users, with $12$ million daily exposed pages and $1$ million Daily Active Users (DAU). During seven days of A/B testing, we collect data and calculate gaps between the considered $4$ models as summarized in Table~\ref{tab:online}. The Deep LTR serves as the baseline. Our proposed RA-EGO method can achieve more than $0.76\%$ Conversion Rate (CR) gain on purchasing  and $0.86\%$ Gross Merchandise Volume (GMV) gain compared to Deep LTR, translating into a significant improvement for a large-scale  E-Commerce platform.

\begin{table}[ht]
\small
\begin{tabular}{l|l|l|l}
\toprule \hline
 \textbf{Models} & \textbf{Update} & \textbf{Imprv on CR} & \textbf{Imprv on GMV} \\ \hline
 Deep LTR & Real-time & $0.0000\%$ & $0.0000\%$ \\ \hline
 RA-RE & No update & $-0.7863\% \pm 0.0076$ & $-0.6859\% \pm 0.0184$ \\ 
 RA-BE & Real-time & $0.5510\% \pm 0.0033$ & $0.3212\% \pm 0.0102$ \\ 
 \textbf{RA-EGO} & \textbf{Daily} & $\mathbf{0.7681\% \pm 0.0058}$ & $\mathbf{0.8602\% \pm 0.0112}$ \\ \hline
\bottomrule
\end{tabular}
\vspace{0.06em}
\caption{Improvement in online A/B tests among RA-EGO, RA-RE, and RA-BE methods compared to Deep LTR.}\label{tab:online}
\end{table}
\vspace{-2em}
The above A/B testing result shows that RA-RE includes many trials that do not choose the proper weights for the RA model. Thus there is a severe decrease in the conversion rate and GMV compared to Deep LTR. On the other hand, RA-BE clearly outperforms Deep LTR, implying that Deep LTR has its limitation. We further examine the strategies chosen by RA-BE and find that RA-BE produces the same permutation as Deep LTR for about 87.7\% pages, implying Deep LTR can have correct predictions in most contexts. Finally, the fact that RA-EGO outperforms RA-BE proves RA-EGO can find better weight combinations beyond the expert-designed strategies. We find that the performance improvement in RA-EGO is mainly brought by the weak sub-models, which otherwise get low attentions in baseline models. We select three sub-models, which ranks items by their prices, sales, and historical click-through rate (CTR), and study the relative improvement in KTD of them in Table~\ref{tab:online2}. 

\begin{table}[h]
\small
\begin{tabular}{l|ll|ll|ll}
\toprule \hline
 \textbf{Models} & \textbf{Price} & \textbf{Imprv} & \textbf{Sale} & \textbf{Imprv} &\textbf{CTR} & \textbf{Imprv}\\ \hline
 Deep LTR & $0.503$ & $+0.00\%$	& $0.437$ & $+0.00\%$ &	$0.413$ & $+0.00\%$ \\ \hline
 RA-RE & $\mathbf{0.500}$ &	$\mathbf{-0.50\%}$ & $\mathbf{0.430}$	&	$\mathbf{-1.79\%}$ & $\mathbf{0.404}$ & $\mathbf{-2.00\%}$ \\ 
 RA-BE & $0.505$ & $+0.46\%$ & $0.440$ & $+0.51\%$ & $0.408$ &	$-1.07\%$ \\ 
 \textbf{RA-EGO} & $0.502$ & $-0.05\%$ & $0.436$ & $-0.34\%$ &	$0.410$ &	$-0.56\%$ \\ \hline
\bottomrule
\end{tabular}
\vspace{0.03em}
\caption{Value and relative improvement of KTD among RA-EGO, RA-RE, and RA-BE methods compared to Deep LTR.}\label{tab:online2}
\end{table}
\vspace{-2em}

Note that a \emph{smaller} KTD value in the Table~\ref{tab:online2} indicates 
a higher order consistency. 
As ranking according to Price or Sale is very different from ranking according to online revenue, there is usually a decrease in the weights of weak sub-models when we optimize the online revenue. In RA-RE, the weights of the weak sub-models are relatively high, and it reasonably leads to a decrease in online revenue. Different from RA-RE, RA-BE optimizes the online revenue, and the weights on Price and Sale decrease. However, RA-EGO produces permutations more 
consistent to the outputs of weak sub-models than Deep LTR, while also provides a better online revenue. This fact further implies RA-EGO may be a better LTE framework than simply applying the deep ensemble learning model.


\section{Conclusion}
We propose the LTE framework for online ranking services and successfully earn considerable online revenue. The LTE framework is scalable to be deployed to improve the revenue of other real-world applications. Our TournamentGreedy is not only a better RA model in classic examinations, but is also the first contextual RA model designed for optimizing the online revenue instead of offline metrics. It needs to serve as a part of the huge online system, and its parameters (weights of permutations) need to be properly chosen by EGO as an RA-EGO framework works. To ensure an RA model can produce satisfying permutations, we emphasize the importance of the expressive power of RA models, and propose weak PO to efficiently estimate the expressive power. RA-EGO is the start of industrial LTE applications. More theories of applied RA models and contextual BBO need to be carefully studied in the future.
 


\bibliographystyle{unsrt}
\balance
\bibliography{sample-base}


\end{document}